%% file: main.tex
\documentclass[letterpaper, 10 pt, conference]{ieeeconf} 
\IEEEoverridecommandlockouts                              
\usepackage{amsmath} 
\usepackage{amssymb}  
\usepackage{graphicx}
\usepackage{color}
\usepackage{stfloats}  
\usepackage{amstext}
\usepackage{transparent}
\usepackage{wasysym}

\title{\LARGE \bf Trajectory Planning and Control for \\ Robotic Magnetic Manipulation } 
\author{Ogulcan Isitman$^{1}$, Gokhan Alcan$^{2}$, Ville Kyrki$^{1}$
\thanks{$^{1}$The authors are with the Department of Electrical Engineering and Automation, School of Electrical Engineering, Aalto University, 00076 Aalto, Finland 
(e-mail: ogulcan.isitman@aalto.fi;  ville.kyrki@aalto.fi)%
}
\thanks{$^{2}$The author is with the Faculty of Engineering and Natural Sciences, Automation Technology and Mechanical Engineering, Tampere University, Finland
(e-mail: gokhan.alcan@tuni.fi)}%
}

\usepackage[colorlinks=false]{hyperref}

\begin{document}
\maketitle
\thispagestyle{empty}
\pagestyle{empty}

\begin{abstract}
Robotic magnetic manipulation offers a minimally invasive approach to gastrointestinal examinations through capsule endoscopy. 
However, controlling such systems using external permanent magnets (EPM) is challenging due to nonlinear magnetic interactions, especially when there are complex navigation requirements such as avoidance of sensitive tissues. 
In this work, we present a novel trajectory planning and control method incorporating dynamics and navigation requirements, using a single EPM fixed to a robotic arm to manipulate an internal permanent magnet (IPM). 
Our approach employs a constrained iterative linear quadratic regulator that considers the dynamics of the IPM to generate optimal trajectories for both the EPM and IPM. 
Extensive simulations and real-world experiments, motivated by capsule endoscopy operations, demonstrate the robustness of the method, showcasing resilience to external disturbances and precise control under varying conditions. 
The experimental results show that the IPM reaches the goal position with a maximum mean error of 0.18 cm and a standard deviation of 0.21 cm.
This work introduces a unified framework for constrained trajectory optimization in magnetic manipulation, directly incorporating both the IPM's dynamics and the EPM's manipulability.
\end{abstract}
 

\section{Introduction} %

Robotic magnetic manipulation has gained significant attention in capsule endoscopy for its potential to provide a non-invasive, painless alternative to traditional endoscopy for diagnosing gastrointestinal diseases \cite{hanscom2022endoscopic}. Magnetically actuated wireless camera capsules, typically equipped with an internal permanent magnet (IPM), not only simplify the internal design but also have the potential to advance capsule endoscopy by enabling active interventions, offering enhanced diagnostic and therapeutic capabilities in procedures \cite{chen2022magnetically}. However, their navigation and control remain mainly dependent on the precise manipulation of an external magnetic field, highlighting the critical role of external magnetic systems in ensuring effective operation.

\begin{figure}[t]
    \centering
    \includegraphics[width=0.4\textwidth]{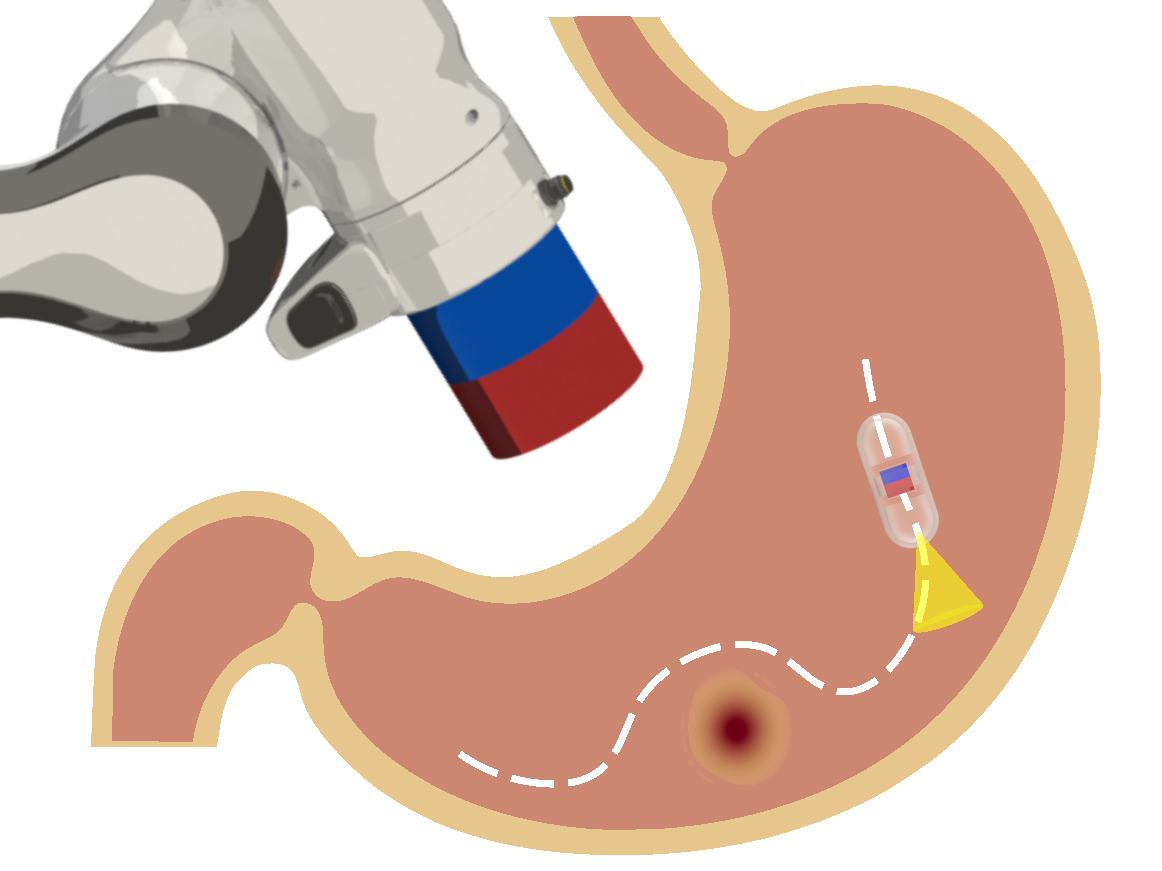}
    \caption{Conceptual design of the external permanent magnet-based robotic capsule endoscopy system. Constrained trajectory optimization accounts for capsule dynamics, enabling precise control and obstacle avoidance. }
    \label{fig:GI}
    \vspace{-15pt}
\end{figure}

Two main methods exist for generating external magnetic fields in such magnetically actuated systems. The first method involves coil-based systems \cite{icsitman2021non, isitman2022simultaneous}, which have proven effective at generating magnetic fields and gradients within a confined workspace  \cite{ongaro2018design,song2022motion}, but they suffer from significant drawbacks, including high power consumption, heat dissipation, and challenges related to their size and weight.
The second method relies on external permanent magnet (EPM) to generate force and torque on the IPM. These magnets can generate strong magnetic fields without the need for an active power source and are typically positioned using a serial robotic arm \cite{taddese2016nonholonomic,pittiglio2019magnetic, xu2022evaluation}, which provides an extended workspace compared to stationary coil-based systems. Despite its advantages, systems with permanent magnets present challenges in precise control due to the highly nonlinear relationship between the magnetic field and the relative position of the EPM to the IPM.

In order to ease this challenge, existing studies with EPMs primarily focus following predefined simple trajectories, which dictate the velocity and pose of the IPM throughout the motion path \cite{mahoney2016five,taddese2016closed}. These approaches have shown promising results in improving control precision and system stability. However, in capsule endoscopy, the operator often needs to perform tasks like monitoring specific regions and navigating constraints such as avoiding sensitive tissues, which require complex maneuvers. These maneuvers cannot be easily achieved using predefined trajectories for systems with EPMs. Automating such manipulation could significantly improve overall system performance and, in turn, diagnostic accuracy, but it remains an open problem.

In this work, we propose a solution to the aforementioned problem by optimizing the trajectories for both the EPM and IPM, taking into account inherent and external constraints. We formulate a constrained iterative linear quadratic regulator (iLQR) for the combined state transitions of the EPM and IPM, incorporating their dynamics and constraints within an augmented Lagrangian framework. The optimized state-input trajectories, along with optimal time-varying local controller gains, are then used for real-time, closed-loop speed control of the joints of the robotic arm that holds the EPM at its end-effector. This formulation allows us to limit the velocity of the IPM and the applied magnetic force while simultaneously maximizing the manipulability of the EPM. To the best of our knowledge, this is the first work to present a constrained trajectory optimization method for magnetic capsule endoscopy that directly considers the IPM's dynamics.

The key contributions of this paper include:
\begin{itemize}
\item A novel formulation for constrained trajectory optimization in magnetic capsule endoscopy, which incorporates the combined dynamics of the EPM and IPM for effective and coordinated control (Fig. \ref{fig:GI}).

\item Introducing the capability to constrain the IPM’s position and velocity, with added obstacle avoidance, while optimizing the EPM's manipulability—thereby enhancing both safety and maneuverability.

\item Developed a simulation environment to test the optimization and control methods, assessing both manipulability and disturbance rejection capabilities to ensure robustness.
\item Performed extensive statistical validation through repeated real-world experiments with optimized trajectories and closed-loop control gains, demonstrating the approach's feasibility in constrained scenarios.

\end{itemize}
\section{Related Work}
Many research efforts have been made to explore and develop closed-loop control architectures for capsule endoscopy, aiming to meet the requirements of clinical applications. These successful efforts can be grouped by their level of autonomy, as defined in the field and described by \cite{martin2020enabling}. The first level is no autonomy, where the operator directly controls the robotic arm to generate a magnetic field that guides the endoscope. This approach is not intuitive due to the highly nonlinear nature of the forces and torques acting on the endoscope. The second level involves the operator controlling the endoscope while the system calculates the necessary movements to achieve the desired motion. This level of autonomy is more intuitive and has been widely accepted, establishing that a minimum of level 2 autonomy is required for the practical use of magnetic endoscopy. In the third level, the system autonomously generates the endoscope's motion based on real-time visual feedback.

To demonstrate the capabilities of systems utilizing level 2 autonomy and to address the nonlinear control challenges inherent in magnetic endoscopy, significant progress has been made in the development of EPM-based systems. For instance, the first 5-DoF control of an endoscopic capsule was achieved using a combination of open-loop orientation control and feedback positioning, enhanced by PID and constrained least squares optimization to follow predefined paths \cite{mahoney2016five}. Similarly, a single EPM and IPM system with closed-loop torque control has been used to levitate and steer the capsule, employing a backstepping-based controller proposed in \cite{pittiglio2019magnetic} to follow a piecewise-constant trajectory for the IPM. Notably, this approach does not account for the system's dynamic equilibrium, allowing velocity and acceleration to be neglected. In \cite{alcaide2023motion}, a set of motion primitives for the EPM was derived to optimize the navigation force applied to the magnetic capsule. This optimization was achieved by minimizing the contact force and explicitly considering the interaction between the capsule and its environment during trajectory generation.

The majority of works falls under level 2 autonomy, focusing on following a predefined trajectory \cite{taddese2016nonholonomic, xu2021reciprocally}. In this context, the trajectory that the IPM follows is already set in advance, dictating the velocity and pose of the IPM throughout the motion path.  However, we hypothesize that enhancing the capsule's control capabilities through trajectory optimization can potentially improve medical imaging quality significantly and, consequently, improve diagnostic accuracy. Improved control allows for smoother, more precise navigation, which is crucial for capturing high-quality images of the gastrointestinal tract. Furthermore, the practicality of the operation would be positively affected if the operator provides only intuitive inputs to the system, such as the desired end position, obstacles to avoid, regions of interest for the camera to focus on, or the desired speed of motion. By reducing the complexity of inputs required from the operator, the system becomes more user-friendly and efficient, potentially bridging the gap towards level 3 autonomy.

A recent attempt at trajectory optimization is presented in \cite{brockdorff2024hybrid}, which proposes a two-stage hybrid trajectory planning method for a dual EPM platform. The first stage involves determining a collision-free path for multiple robotically actuated permanent magnets, ensuring obstacle-free motion while accounting for magnetic interactions between the two EPMs for safety. The second stage focuses on planning in magnetic space to achieve smooth transitions in magnetic actuation. This method operates in the operational space, requiring an additional inverse kinematics step to generate joint space trajectories. Moreover, this approach primarily aims at generating obstacle-free EPM trajectories to steer a soft robot along a predefined path in magnetic space without considering the dynamics. Another recent work proposes a motion control method for a dual EPM system that optimizes the EPM pose to generate tilting and rotational motions without causing undesired translational motion of the IPM \cite{bae2024optimal}. However, the proposed method does not address the dynamic motion of the EPM or trajectory planning for the IPM.

In contrast, our method directly incorporates the dynamics of the IPM and operates in the joint space, eliminating the need for additional inverse kinematics computations. By doing so, we address the limitations of previous methods by providing a unified framework that optimizes the trajectory of both the EPM and the IPM, ensuring safe, efficient, and precise navigation of the capsule endoscope. We have assessed our approach's accuracy, efficiency, and robustness through both simulation analyses under various external disturbances and experimental evaluations, thereby ensuring the method's comprehensive reliability.

\begin{figure*}[ht]
    \centering
    \def\svgwidth{0.9\linewidth}
    {
    \fontsize{9}{9}
    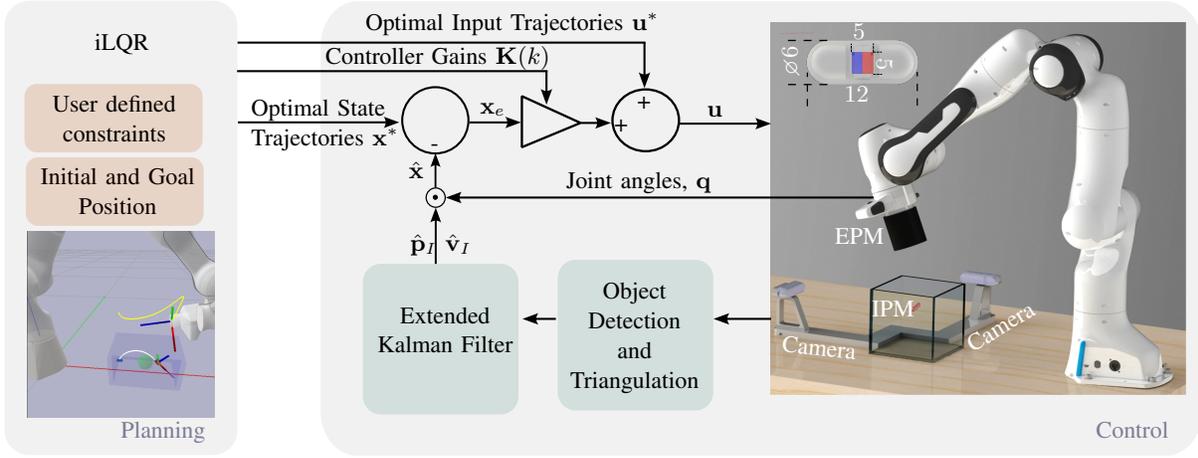
    }
    \caption{Block diagram of the planning and control phases for robotic magnetic manipulation. The planning phase (left) utilizes iLQR to generate optimal state trajectories \( \mathbf{x}^* \), input trajectories \( \mathbf{u}^* \), and time-varying optimal controller gains \( \mathbf{K}(k) \), based on user-defined constraints and initial/goal positions. In the control phase (right), joint angles \( \mathbf{q} \) are measured from the robotic arm encoders, and the 3D position of the IPM is obtained through object detection and triangulation using two orthogonal cameras. The Extended Kalman Filter then estimates the IPM's position \( \hat{\mathbf{p}}_I \) and velocity \( \hat{\mathbf{v}}_I \), which are used to update the control input \( \mathbf{u} \) and follow the optimal trajectories. The system includes the EPM for external magnetic manipulation and the IPM within the targeted environment, with the IPM dimensions shown in the figure and given in millimeters.}
    \label{fig:block_diagram}
    \vspace{-10pt}
\end{figure*}

\section{Proposed Method}
\subsection{Problem statement}
We consider a scenario where a  single EPM attached to a robotic arm is used to navigate an IPM inside the human body for a video endoscopy procedure.
We assume that the EPM is positioned with a robotic arm with at least 5 DoF and the magnetization of the IPM is constant and quickly aligns with the magnetic field generated by the EPM. The IPM is manipulated inside a fluid-filled environment and the 3D position
of the capsule is observable. 

We formulate trajectory optimization as a constrained optimal control problem where the objective is to minimize a cost function while satisfying the system dynamics and constraints. For a given initial state $\mathbf{x}_0$, goal state $\mathbf{x}_g$ and a time horizon $N$, the discrete-time constrained optimal control problem can be formulated as follows
\begin{equation}
    \begin{aligned}
        & \underset{u_{0:N-1}}{\text{minimize}} && \ell_N(\mathbf{x}_N) +\sum_{k=0}^{N-1} \ell_k(\mathbf{x}_k, \mathbf{u}_k, \Delta t)  \\
        & \text{subject to} && \mathbf{x}_{k+1} = f(\mathbf{x}_k, \mathbf{u}_k, \Delta t), \\
        &&& g(\mathbf{x}_k, \mathbf{u}_k) \leq 0, \quad h(\mathbf{x}_k, \mathbf{u}_k) = 0, \\
        &&& \mathbf{u}_{\text{min}} \leq \mathbf{u}_k \leq \mathbf{u}_{\text{max}}, \\
        & \text{given} && \mathbf{x}_0
    \end{aligned}
    \label{eqn:opt}
\end{equation}
where $\ell_N$ and $\ell_k$ are the terminal and running costs, $g(.)$ and $h(.)$ are the inequality and equality constraints, respectively.
The combined state transition function is represented as $f(\cdot)$, which includes discretized system dynamics and the forward kinematics of the robotic arm. This function defines the interaction between the EPM and IPM, where $\mathbf{x} \in \mathbb{R}^{13}$ denotes the state, and $\mathbf{u} \in \mathbb{R}^{7}$ denotes the controls. It is assumed that all state variables are observable for the system.

\subsection{System Modelling}
We defined the states and controls as:
\begin{equation}
\label{eq-state-input}
    \begin{aligned}
        \mathbf{x} & = [\mathbf{p}_{I}, \mathbf{v}_{I}, \mathbf{q}] \quad , \quad
        \mathbf{u}= [\omega_1, \ldots, \omega_7],
    \end{aligned}
\end{equation}
where $\mathbf{p}_{I}, \mathbf{v}_{I} \in \mathbb{R}^3$ represent the IPM's position and velocity, respectively, and $\mathbf{q} \in \mathbb{R}^7$ represents the joint angles of the robotic arm that actuates the EPM, while the control vector represents the joint velocities ($\omega_1$, ..., $\omega_7$) of the robotic arm.

Selecting joint angles within the states provides multiple advantages, including the ability to impose constraints in joint space, apply restrictions on the EPM and IPM, and define obstacles within the operational space. Since the optimization accounts for the robot's kinematics, the resulting trajectory is guaranteed to be feasible for the robotic actuator and eliminates the need for an additional inverse kinematics solver. The EPM's position and orientation are determined using standard forward kinematics calculations for a robotic arm \cite{craig2006introduction} as:
\begin{equation}
    \label{eq-fwd}
    \begin{bmatrix} \mathbf{R}_E & \mathbf{p}_E \\ 0 & 1 \end{bmatrix} = \texttt{fwd}(\text{DH}, \mathbf{q}) 
\end{equation}
where the $\texttt{fwd}$ function accounts for the Denavit-Hartenberg (DH) parameters and joint angles ($\mathbf{q}$). Here, \( \mathbf{p}_E \) denotes the position of the EPM, while \( \mathbf{R}_E \) represents its orientation relative to the base frame. The dynamics of the robot arm are not considered explicitly because it is assumed that a low-level arm controller is able to control the arm within specified velocity and acceleration limits that are specified as constraints in the optimization problem (\ref{eqn:opt}).

The magnetic interaction between the EPM and the IPM can be defined using the point dipole
field approximation \cite{abbott2020magnetic}, where the dipole moment of both the EPM and 
the IPM is parallel to their principal axes and is denoted by $\mathbf{m}_{E}$ and 
$\mathbf{m}_{I} \in R^3$ , respectively.

The magnetic field generated by the EPM at the position of the IPM, where the positions are defined at the magnet centers, with the distance between them given by $\mathbf{p}$$=$$\mathbf{p}_I$$-$$\mathbf{p}_E$, is expressed as follows:

\begin{equation}
    \mathbf{b}(\mathbf{p},\mathbf{m}_{E}) = \frac{\mu_0}{4\pi \lVert \mathbf{p} \rVert^3} \left( 3\mathbf{\hat{p}}\mathbf{\hat{p}}^T - \mathbf{I}_3 \right) \mathbf{m}_E
\end{equation}
where $\mu_0$ is a constant for the free space permeability,  $\mathbf{\hat{p}} = \mathbf{p} / \lVert \mathbf{p} \rVert$ is the unit vector in the direction of $\mathbf{p}$ and $\mathbf{I}_3 \in R^{3 \times 3 }$ is the identity matrix. 

Under the applied magnetic field, the magnetic dipole experiences a force and a torque to minimize the magnetic potential energy.
The force and torque acting on the IPM can be calculated as
\begin{equation}
    \begin{aligned}
        \mathbf{f}_m &= (\mathbf{m}_I \cdot \nabla)\mathbf{b} \\
        \mathbf{\tau}_m &= \mathbf{m}_I \times \mathbf{b}
    \end{aligned}
\end{equation}
where $\nabla$ is the gradient operator. Therefore the gradient of the magnetic field can be calculated as
\begin{multline}
    \nabla \mathbf{b}(\mathbf{p},\mathbf{m}_E) =  \frac{3\mu_0}{4\pi \lVert \mathbf{p} \rVert^4} \\ \left( \mathbf{m}_E \mathbf{\hat{p}}^T + \mathbf{\hat{p}}^T\mathbf{m}_E^T  + \left(\mathbf{\hat{p}}^T \mathbf{m}_E\right) \left( I_3 - 5\mathbf{m}_E\mathbf{m}_E^T \right) \right) .
\end{multline}

The IPM magnetization is assumed to be constant and will quickly align with the magnetic field generated by the EPM.
This assumption holds under the condition that the IPM moves with small accelerations and is not in contact with other objects.
It implies that there is no need for orientation feedback to control the orientation, 
and it can further simplify the force calculations where $ \hat{\mathbf{b}} \approx \hat{\mathbf{m}}_I $ to
\begin{equation}
    \mathbf{f}_m \approx \lVert \mathbf{m}_I  \rVert(\hat{\mathbf{b}} \cdot \nabla)\mathbf{b}.
\end{equation}
This can be further expanded and rearranged as
\begin{multline}
    \label{eq-mag-force}
     \mathbf{f}_m(\mathbf{p},\mathbf{m}_{E},\mathbf{m}_{I}) \approx  \frac{3\mu_0  \lVert \mathbf{m}_{E} \rVert \lVert \mathbf{m}_{I} \rVert}{4\pi \lVert \mathbf{p} \rVert^4 \left( \rVert 3\mathbf{\hat{p}}\mathbf{\hat{p}}^T - I_3 \right)\hat{\mathbf{m}}_{E}\rVert} \\ \left( \hat{\mathbf{m}}_{E}\hat{\mathbf{m}}_{E}^T -\left( 1 + 4(\hat{\mathbf{m}}_{E}^T \mathbf{\hat{p}})^2\right) I_3 \right)\mathbf{\hat{p}} 
\end{multline}

Finally, the acceleration of the IPM is calculated as: 
\begin{equation}
    \frac{\delta \mathbf{v}_I}{\delta t} = \frac{1}{m_{\text{IPM}}} \left( \mathbf{f}_m + \mathbf{f}_w - \mathbf{f}_d \right)
\end{equation}
where $m_{\text{IPM}}$ is the mass of the IPM, $\mathbf{f}_m$ is the magnetic force as obtained in (\ref{eq-mag-force}), $\mathbf{f}_w$ denotes the effective weight force incorporating both buoyancy and gravity, and $\mathbf{f}_d$ is the drag force calculated as $\mathbf{f}_d=C_d \mathbf{v}_I^2$ with a drag coefficient $C_d$. 

\subsection{Trajectory Planning and Control}
The proposed trajectory planning and control method for magnetic capsule endoscopy is presented in Fig. \ref{fig:block_diagram}. During the offline planning phase, optimal trajectories for both the EPM and IPM are calculated, along with time-varying feedback gains for a state feedback controller, while incorporating user-defined constraints. 

The inequality constraints, $g(\mathbf{x}_k, \mathbf{u}_k)$, are formulated as:
\begin{equation}
\label{eq-ineq-const}
\begin{aligned}
    \text{Joint angle limits:} & \quad q_{min} \leq \mathbf{q}_k \leq q_{max} \\ 
    \text{Joint velocity limits:} & \quad u_{min} \leq \mathbf{u}_k \leq u_{max} \\ 
    \text{IPM velocity limits:} & \quad v_{I,min} \leq \mathbf{v}_{I_k} \leq v_{I,max} \\ 
    \text{EPM position limit:} & \quad p_{E,min} \leq \mathbf{p}_{E_k} \\ 
    \text{Obstacle Avoidance:} & \quad d_{obs} = \| \mathbf{p}_{I_k} - \mathbf{p}_{obs} \| \\ 
     & \quad d(\mathbf{p}_{I_k}) = d_{obs} - r_{obs} - \epsilon \\
     & \quad d(\mathbf{p}_{I_k}) \geq 0
\end{aligned}
\end{equation}
for all $k$$=$$0,...,N$. Here, \( q_{min} \) and \( q_{max} \) denote the minimum and maximum joint angles, defining the allowable range of motion for each joint. \( u_{min} \) and \( u_{max} \) set the limits for joint velocities, while \( v_{I,\min} \) and \( v_{I,\max} \) constrain the IPM’s velocity. \( p_{E,\min} \) represents the lower bound for the EPM's position, ensuring it stays within a safe operational range. \( d_{obs} \) is the Euclidean distance between the IPM's position \( \mathbf{p}_{I} \) and the obstacle's position \( \mathbf{p}_{obs} \), while \( d(\mathbf{p}_I) \) is the adjusted distance, accounting for the obstacle radius \( r_{obs} \) and a safety margin \( \epsilon \). The condition \( d(\mathbf{p}_I) \geq 0 \) ensures a safe distance is maintained from the obstacle.

The equality constraint, $h(\mathbf{x}_k, \mathbf{u}_k)$, is formulated as:
\begin{equation}
\label{eq-eq-const}
\begin{aligned}
    \text{IPM Orientation: } &\mathbf{b}(\mathbf{p},\mathbf{m}_{E}) =  \mathbf{r}\\ 
\end{aligned}
\end{equation}
where \( \mathbf{r} \in \mathbb{R}^3 \) is the target orientation vector. This vector specifies the desired alignment of the IPM, with its x-axis oriented along the direction of its magnetization $\hat{\mathbf{m}}_I$.

We proposed an iterative LQR algorithm to solve the trajectory optimization problem defined in (\ref{eqn:opt}), as it is computationally efficient for such problems involving high-dimensional state spaces and long horizons. Additionally, it inherently provides locally optimal time-varying controller gains that can be utilized in closed-loop, real-time control. To incorporate the constraints formulated in (\ref{eq-ineq-const}) and (\ref{eq-eq-const}), we employed the augmented Lagrangian method \cite{howell2019altro}. The augmented Lagrangian function is defined as follows:
\begin{equation}
\begin{aligned}
\mathcal{L}_A&= \mathcal{L}_N(\mathbf{x}_N) + \sum_{k=0}^{N-1} \mathcal{L}_k(\mathbf{x}_k, \mathbf{u}_k) \\
\mathcal{L}_N(\mathbf{x}_N) &= \ell_N(\mathbf{x}_N) + (\lambda + \frac{1}{2}\mathbf{g}(\mathbf{x}_N)\mathcal{I}_\mu)^T\mathbf{g}(\mathbf{x}_N) \\
\mathcal{L}_k(\mathbf{x}_k,\mathbf{u}_k) &= \ell_k(\mathbf{x}_k,\mathbf{u}_k) + (\lambda + \frac{1}{2}\mathbf{g}(\mathbf{x}_k,\mathbf{u}_k)I_\mu)^T\mathbf{g}(\mathbf{x}_k,\mathbf{u}_k)
\end{aligned}
\end{equation}
where $\lambda$ are the Lagrangian multipliers, $I_\mu$ is a diagonal matrix composed by penalty parameter $\mu$, $\mathbf{g}(\cdot)$ is the vector of all inequality and equality constraints. Given the augmented Lagrangian cost functions for a fixed values of $\lambda$ and $\mu$, the cost-to-go ($V$) and action-value ($Q$) functions are
\begin{equation}
\label{eq:Q_and_V}
\begin{aligned}
    V_k(\mathbf{x}_k) & =\underset{\mathbf{u}_k}{\min}\{\mathcal{L}_k(\mathbf{x}_k,\mathbf{u}_k)+V_{k+1}(f(\mathbf{x}_k,\mathbf{u}_k))\} \\
    &=\underset{\mathbf{u}_k}{\min}\;Q(\mathbf{x}_k,\mathbf{u}_k).
\end{aligned}
\end{equation}
where the action-value function $Q$ is approximated up to second order as:

\begin{equation}
\label{quad}
\begin{aligned}
    Q(\mathbf{x} + \delta\mathbf{x}, \mathbf{u} + \delta\mathbf{u}) &\approx Q(\mathbf{x}, \mathbf{u}) + Q_\mathbf{x}^\top \delta \mathbf{x} + Q_\mathbf{u}^\top \delta \mathbf{u} \\
    &\quad + \frac{1}{2} (\delta\mathbf{x} ^\top Q_{\mathbf{xx}} \delta \mathbf{x} + \delta\mathbf{u}^\top Q_{\mathbf{uu}} \delta \mathbf{u}) \\
    &\quad + \delta \mathbf{x}^\top Q_{\mathbf{xu}} \delta\mathbf{u}.
\end{aligned}
\end{equation}
The derivatives of $Q$ function are calculated as
\begin{equation}
\begin{aligned}
    Q_\mathbf{x} &= \mathcal{L}_\mathbf{x} + f_{\mathbf{x}}^\top V_\mathbf{x}^{\prime}, &Q_\mathbf{xx} &= \mathcal{L}_\mathbf{xx} + f_{\mathbf{x}}^\top V_\mathbf{xx}^{\prime}f_{\mathbf{x}}, \\
    Q_\mathbf{u} &= \mathcal{L}_\mathbf{u} + f_{\mathbf{u}}^\top V_\mathbf{x}^{\prime}, &Q_\mathbf{uu} &= \mathcal{L}_\mathbf{uu} + f_{\mathbf{u}}^\top V_\mathbf{xx}^{\prime}f_{\mathbf{u}}, \\
    & &Q_\mathbf{ux} &= \mathcal{L}_\mathbf{ux} + f_{\mathbf{u}}^\top V_\mathbf{xx}^{\prime}f_{\mathbf{x}} 
\end{aligned}
\end{equation}
where $V_\mathbf{xx}^{\prime}$ and $V_\mathbf{x}^{\prime}$ are the Hessian and gradient of the cost-to-go at time step $k+1$, respectively. 

Minimizing (\ref{quad}) with respect to $\delta\mathbf{u}$ results in an affine controller
\begin{equation}
\begin{aligned}
    \delta_\mathbf{u}^* &= -Q_{\mathbf{uu}}^{-1}(Q_{\mathbf{ux}}\delta\mathbf{x} + Q_\mathbf{u}) \triangleq \mathbf{K} \delta \mathbf{x} + \mathbf{d}.
\end{aligned}
\end{equation}

Subsequently, the optimal controller gains are employed in the calculations of the derivatives of $V$ function as
\begin{equation}
\label{eq:V-derivatives}
    \begin{aligned}
        V_{\mathbf{x}} & = Q_{\mathbf{x}}+\mathbf{K} Q_{\mathbf{u}} + \mathbf{K}^\top Q_{\mathbf{uu}}\mathbf{d}+Q_{\mathbf{ux}}^\top \mathbf{d},\\
        V_{\mathbf{xx}} & = Q_{\mathbf{xx}}+\mathbf{K}^\top Q_{\mathbf{uu}}\mathbf{K}+\mathbf{K}^\top Q_{\mathbf{ux}}+Q_{\mathbf{ux}}^\top \mathbf{K}.
    \end{aligned}
\end{equation}
In iLQR algorithm, the objective function is optimized iteratively by following backward and forward passes \cite{alcan2022differential}. In the backward pass, all local optimal controller gains ($\mathbf{K,d}$) are sequentially calculated in reverse, starting from the final state, where $V_N(\mathbf{x}_N)=\mathcal{L}_N(\mathbf{x}_N)$ and progressing back to the initial state. In the forward pass, those controller gains are employed to update the control sequence starting from the initial state. After convergence is achieved, $\lambda$ and $\mu$ parameters are updated as described in \cite{alcan2023constrained}.

After planning is completed, optimal state and input trajectories as well as control gains are employed in closed-loop control (Fig. \ref{fig:block_diagram}). In the control phase, the 3D position of the IPM is determined using an object detection algorithm with two orthogonal cameras and the triangulation method \cite{hartley2003multiple}. Extended Kalman Filter (EKF) \cite{lefebvre2004kalman} algorithm is utilized to estimate the position and velocity of the IPM, ($\mathbf{\hat{p}}_I, \mathbf{\hat{v}}_I$). The joint angles of the robotic arm ($\mathbf{q}$) are directly measured from the encoders, and these signals are concatenated to form the measurement for the state vector ($\mathbf{\hat{x}}$) defined in (\ref{eq-state-input}). Lastly, the state feedback control is applied as:
\begin{equation}
\label{eq-closed-loop}
\begin{aligned}
    \mathbf{\bar{u}}_k = \mathbf{u}^*_k + \mathbf{K}_k(\mathbf{\hat{x}}_k - \mathbf{x}^*_k)
\end{aligned}
\end{equation}
where $\mathbf{u}^*_k \in \mathbb{R}^{7}$ and $\mathbf{K}_k \in \mathbb{R}^{7 \times 13}$ represent the optimal control input and the time-varying feedback gain matrix at each time step $k$, respectively. The variables $\mathbf{x}^*_k$ and $\mathbf{\hat{x}}_k \in \mathbb{R}^{13}$ are the optimal states obtained through iLQR and the state measurements, respectively.

\section{Experiments}
\label{sec:exp}
To evaluate the performance of the proposed method for capsule endoscopy, our experiments investigate: 1) the adherence of the optimal trajectory to constraints, 2) the robustness and performance of trajectory following in both simulations and real-world scenarios, and 3) the ability to navigate the IPM within potential obstacles in the environment, such as polyps or tumors, to minimize interaction and enhance screening effectiveness.

\subsection{Experimental Setup}

The proposed method was validated using the capsule IPM and a single moving EPM system (Figure \ref{fig:block_diagram}). A 7-DoF Franka Emika Panda serial robotic arm was used to position the EPM, which consisted of a cylindrical neodymium magnet with dimensions of 45 mm in diameter and 30 mm in height, a dipole moment of 51.25 A·m², and a grade of N45.
The IPM was a 3D-printed, air-filled capsule housing a cubic neodymium magnet with dimensions of 5 mm × 5 mm × 5 mm, a dipole moment of 0.142 A·m², and a grade of N50.

The IPM capsule is navigated inside a water-filled tank with a volume of $15 \times 15 \times 15  \text{ cm}^3$.
Two orthogonal cameras (Intel RealSense D435) are employed to track the IPM's position in 3D, 
with a sampling rate of 50 Hz. The capsule detection is achieved through the YOLO algorithm \cite{JocherUltralyticsYOLO2023}, executed on a NVIDIA Titan Xp GPU.

The system parameters were identified through experimental estimation, consisting of a drag coefficient ($C_d$) of 0.77, an effective weight force ($\mathbf{f}_w$) of 0.69 mN, and an IPM mass of 8.1 g.

The simulation environment is developed to evaluate optimal constrained trajectory performance, mimicking the real system using estimated parameters and the proposed control method. It includes models of the robotic arm and 3D representations of the EPM and IPM, with physical properties, while computing the necessary Hessian and gradients via automatic differentiation using the \texttt{jax} library \cite{jax2018github}.

\section{Results}
We evaluated the performance of the optimal trajectory planning method in the presence of nonlinear constraints, including IPM and EPM configurations, obstacle avoidance, joint position and velocity, and the manipulability of the actuator robotic arm. Moreover, we evaluate the disturbance rejection capabilities and repeatability of the method in real world setting \footnote{The videos of the experiments can be found on the project website: \href{https://sites.google.com/view/robomagce}{\texttt{https://sites.google.com/view/robomagce}}}. 

\subsection{Simulation Results}
We present a case where the objective is to determine the optimal trajectory for the IPM, ensuring it avoids collisions with a spherical obstacle (e.g., stomach polyps) while maintaining the desired orientation throughout the path, thereby enhancing visual examination. Additionally, we aim to maximize the manipulability of the EPM during the entire trajectory to improve dexterity.

\begin{figure*}[t]
    \def\svgwidth{0.8\linewidth}
    \centering
    {
    \fontsize{9}{9}
    \includegraphics[width=\textwidth]{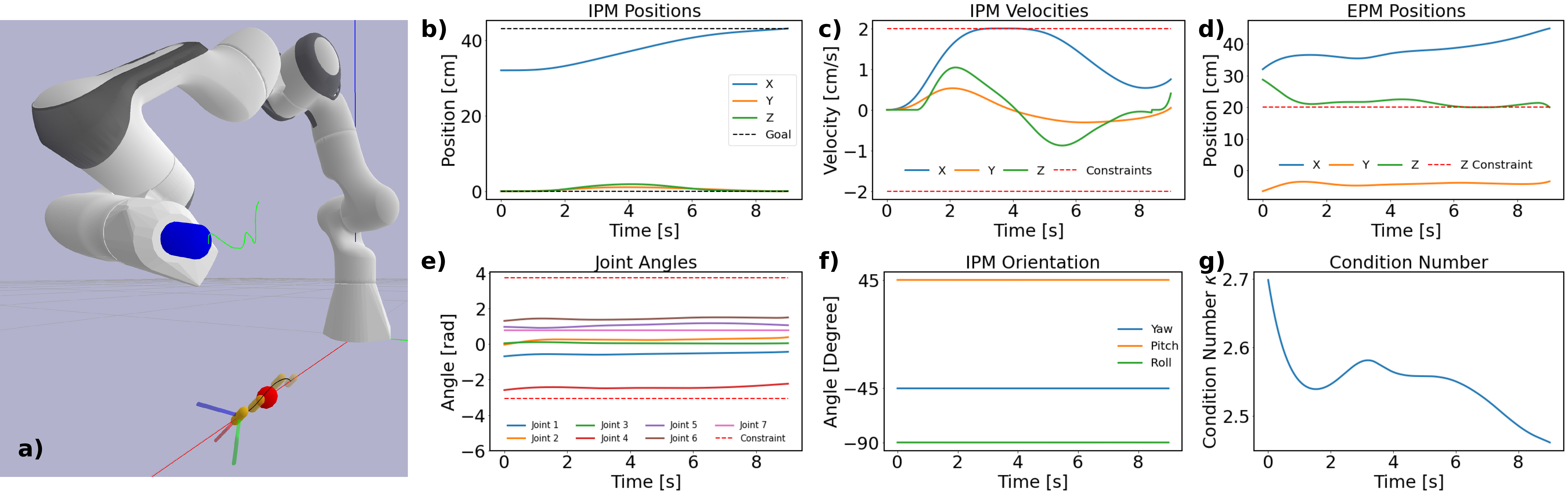}
    }
    \vspace{-10pt}
\caption{(a) Simulation case visualization. (b) Time series data of the optimal IPM trajectory. (c) IPM velocities along the trajectory. (d) Time series data of the optimal EPM trajectory. (e) Joint angles of the robotic arm. (f) Time series data of the IPM orientation. (g) Condition number (\(\kappa\)) analysis of the robotic arm’s manipulability.}
    \label{fig:sim}
    \vspace{-10pt}
\end{figure*}

The optimal trajectories of both the IPM and EPM are plotted, along with the initial, final, and two intermediate poses of the IPM throughout the optimal trajectory, in Fig. \ref{fig:sim} (a). The IPM successfully reaches the final position without colliding to the obstacle while keeping its orientation constant. The positional trajectories of the IPM and EPM over time are presented in Fig. \ref{fig:sim}(b) and (d), showing their paths as IPM progress toward the goal position. As observed from Fig. \ref{fig:sim} (d), achieving a relatively smooth and simple IPM trajectory requires a non-trivial dynamic motion for EPM, highlighting the significance of combined trajectory planning.

The adherence of the optimal trajectory to the constraints is evident in Fig. \ref{fig:sim}(c), (d), and (e). In Fig. \ref{fig:sim}(c), the IPM velocity is constrained to 20 cm/s. The maximum velocity is observed along the X-axis, where it is capped by the given constraint. The EPM's position is also constrained in the Z direction, Fig. \ref{fig:sim}(d), to prevent it from colliding with the tank inside which the IPM is operating. This constraint is crucial for ensuring patient safety and avoiding unintended interactions with the robotic arm.
To ensure that the optimal trajectory is feasible for execution within the robotic arm's physical limits, the joint angles are also constrained within the permissible range, as shown in Fig. \ref{fig:sim}(e). The position of Joint 7 remains constant throughout the motion, which is the expected outcome due to the placement of the EPM on the robotic arm. In this configuration, the magnetization is parallel to the joint axis, meaning that rotation around the magnetization axis does not generate any force or torque on the IPM.

The IPM orientation is held constant, as depicted in Fig. \ref{fig:sim} (f), during the entire manipulation, with the initial orientation selected arbitrarily. On the other hand, the robot's initial configuration is calculated to generate zero net force on the IPM while producing the desired magnetic field direction, which is aligned with the desired orientation at the given IPM position.

Furthermore, the manipulability of the robotic arm is maximized by penalizing the high condition number ($\kappa$) \cite{salisbury1982articulated} in the cost functions. Typically, $\kappa$ ranges from $1$ to $\infty$, with a value close to 1 indicating a well-conditioned Jacobian at that point. In Fig. \ref{fig:sim}(g), even though the $\kappa$ increases momentarily between 2 and 4 seconds as the IPM maneuvers around the obstacle, owing to the higher priority of obstacle avoidance, it remains lower than the initial position and quickly reduces after passing the obstacle.

\begin{figure}[t]
    \centering
    \includegraphics[width=1\linewidth]{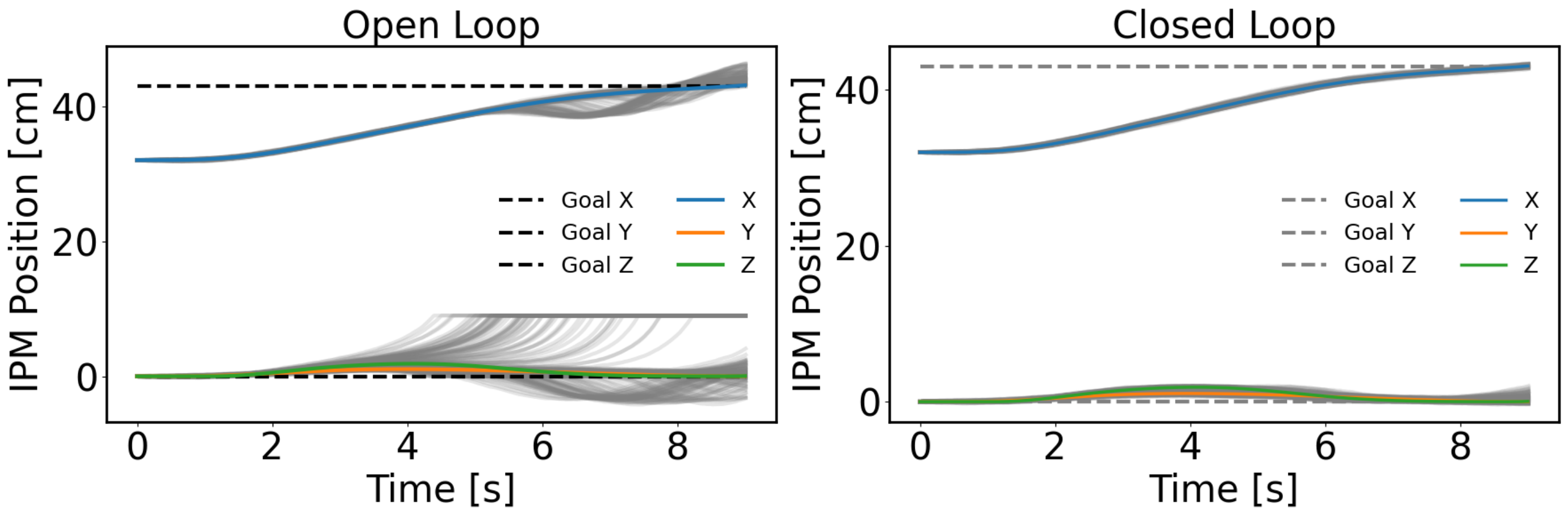}
    \caption{Comparison of open-loop and closed-loop IPM position tracking under Gaussian noise. In the open-loop (left), significant deviations from the desired trajectories (solid colored lines) are observed, with gray lines indicating variability across trials. In the closed-loop (right), the controller reduces deviations, closely following the desired trajectories. Dashed lines represent goal positions, solid colored lines are optimal/desired trajectories, and gray lines show variations across multiple trials.}
    \label{fig:CL}
    \vspace{-10pt}
\end{figure}

To test the disturbance rejection capability of the proposed method, we perturbed the position measurements of the IPM with zero-mean Gaussian noise with a variance of $10^{-2}$. Simulations were repeated 100 times. In open-loop execution of optimal inputs ($\mathbf{u}^*$), IPM positions deviated from the optimal trajectory after 4 seconds. On the other hand, closed-loop application of state feedback control described in (\ref{eq-closed-loop}) successfully rejected the disturbances and kept the trajectory along the optimal one (Fig. \ref{fig:CL}).

\subsection{Real World Results}

We further demonstrated the performance of our method in repetitive real-world settings, considering a similar scenario performed in simulation where the initial and final positions of the IPM are given. The goal is to find the optimal trajectory that avoids a spherical virtual obstacle while adhering to the defined IPM and EPM position and velocity constraints.

\begin{figure}[t]
    \centering
    \def\svgwidth{0.8\linewidth}
    {
    \fontsize{8}{8}
    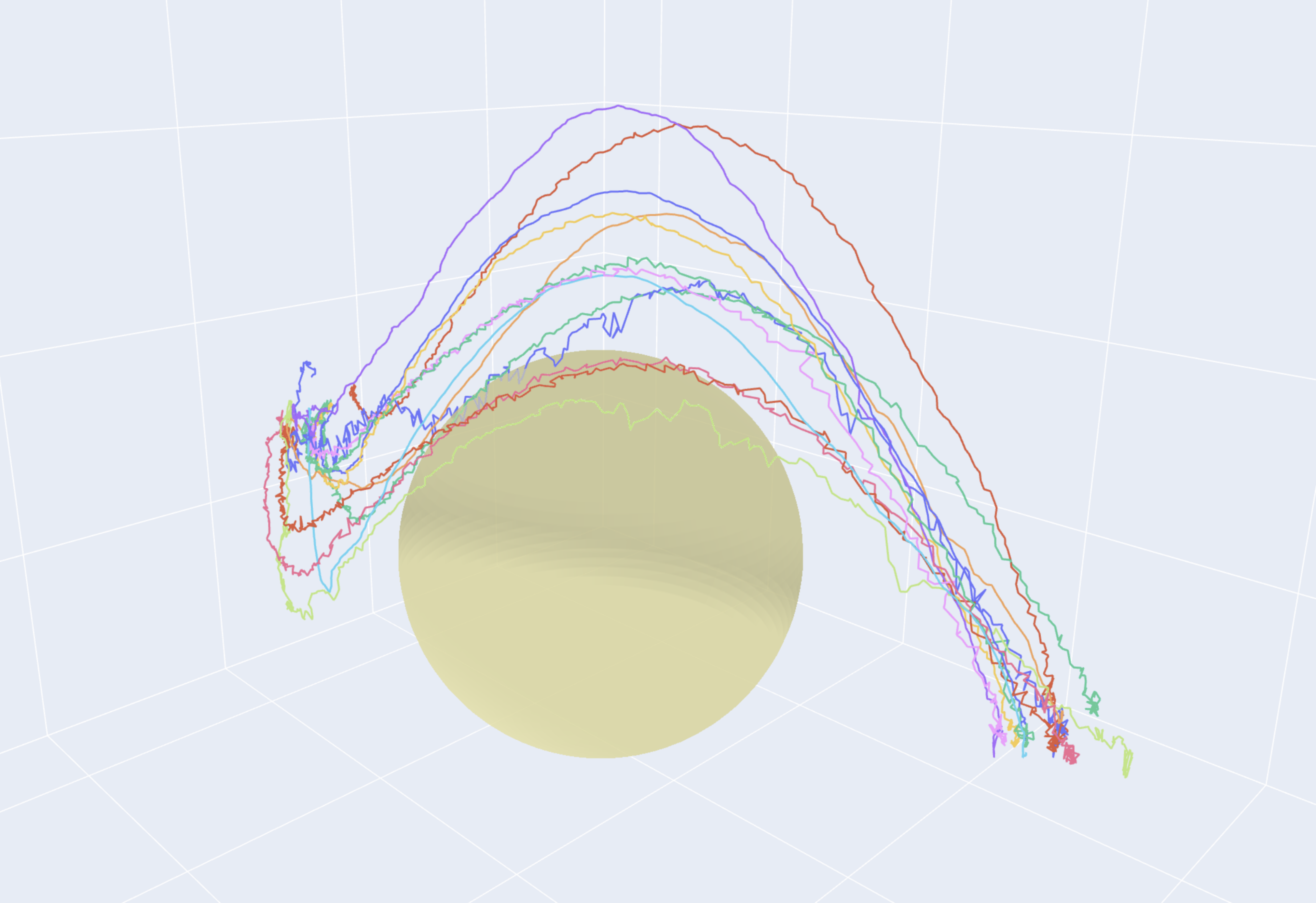
    }
    \caption{Real-world experimental results showing repetitive IPM trajectories in 3D space under the given constraints, including obstacle avoidance (yellow sphere). Each colored line represents an individual trial of the IPM navigating around the obstacle, demonstrating the system's ability to consistently follow the desired path while respecting imposed constraints.}
    \label{fig:3D}
    \vspace{-10pt}
\end{figure}

In Fig. \ref{fig:3D}, the real-world experimental IPM trajectories are presented for 13 repetitions, with a virtual obstacle placed between the initial and goal positions. Before executing each experiment, the IPM is moved to a stable floating point to achieve the same initial position specified in Table \ref{tab:ipm_stats}, with zero initial velocity. The initial position errors in both the X and Y axes stem from the stabilization process of the IPM in the floating position. A PID controller is used for stabilization in the Z-axis, while an open-loop controller is employed for the X and Y axes. As a result, the Z-axis shows almost perfect alignment with the planned trajectory under initial conditions for all repetitions, as can be observed in the deviation band of the IPM position trajectory. Once the IPM reaches and stabilizes at the given initial Z position, the PID controller is deactivated, and the system switches to the proposed iLQR-based controller.
 
\begin{figure*}[t]
    \centering
    \def\svgwidth{0.96\linewidth}
    {
    \fontsize{7}{7}
    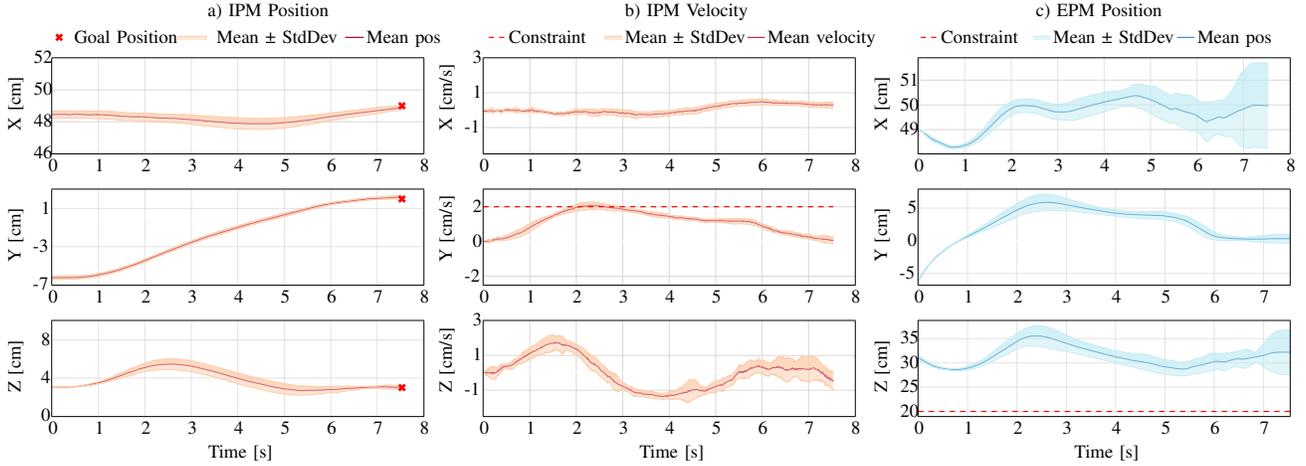
    }
    \vspace{-8pt}
    \caption{Experimental results showing the time evolution of IPM position, IPM velocity, and EPM position along the X, Y, and Z axes under closed-loop control. The left column (a) shows the IPM position, with mean position (solid line) and standard deviation (shaded area) around the mean, alongside goal positions (red crosses). The middle column (b) displays IPM velocity, with mean velocity and standard deviation, and includes velocity constraints (dashed red line) in the Y-axis plot. The right column (c) represents EPM position with mean position (solid line) and standard deviation (shaded area) around the mean, as well as position constraints in the Z-axis plot (dashed red line). This figure demonstrates the tracking accuracy and stability of both IPM and EPM under the control framework.}
    \label{fig:exp_results}
\end{figure*}

IPM positions over time are presented in Fig. \ref{fig:exp_results} (a), showing the mean position with standard deviations to evaluate repeatability. For the X-axis, while the initial and goal positions are the same, small variations arise due to initial condition errors and model inaccuracies. However, the mean values approach the goal position by the final time, where the deviations become minimal relative to the rest of the trajectory.

\renewcommand{\arraystretch}{1.2} 

\begin{table}[b]
\vspace{-8pt}
\centering
\caption{Experimental IPM Position and Velocity Statistics}
\resizebox{\columnwidth}{!}{%
\begin{tabular}{lccc}
\hline
\textbf{Parameter} & \textbf{X} & \textbf{Y} & \textbf{Z} \\
\hline
Initial Position [cm]        & 49.0 & -6.0 & 3.0 \\
Goal Position [cm]           & 49.0 & 2.0  & 3.0 \\
Mean Position at Final [cm]  & 48.870 & 2.182  & 2.964 \\
Std. Dev. Position at Final [cm]   & 0.147  & 0.207  & 0.217 \\
Mean Velocity at Final [cm/s] & 0.281  & 0.068  & -0.455 \\
Std. Dev. Velocity at Final [cm/s]  & 0.171  & 0.214  & 0.502 \\
\hline
\end{tabular}%
}
\label{tab:ipm_stats}
\end{table}

For the Y-axis, the IPM position deviation is small around the mean value for the repetitions, and the final mean positions and their standard deviations are given in Table \ref{tab:ipm_stats}. Conversely, for the Z-axis, an increase in the deviation band is observed between 2 and 5 seconds, where the IPM avoids obstacles. However, these discrepancies are quickly compensated for, allowing the trajectory to reach the goal position by the final time. The observed differences, particularly in the Z-axis, are largely due to modeling inaccuracies. The magnetic force model exhibits an inverse fourth-order relationship with the distance between the EPM and IPM, meaning even small measurement errors can lead to significant variations in the magnetic force.

Fig. \ref{fig:exp_results} (b) presents the IPM velocities over time in the X, Y, and Z axes. On the Y-axis, the imposed velocity constraint is clearly evident. The mean velocity remains within the 2 cm/s limit throughout the entire duration, and the standard deviations are relatively small, indicating that the system consistently adheres to the velocity restriction. For the X and Z axes, the mean velocities exhibit fluctuations around zero, with variations primarily occurring in the early and middle parts of the trajectory. 
The optimization problem also aims to minimize terminal velocity to achieve stability at the goal position. As indicated by the mean velocities at the final time, the velocities for each axis approach zero.

EPM position trajectories, with their corresponding statistics, are also presented in Fig. \ref{fig:exp_results} (c). Unlike the IPM positions, the EPM positions tend to diverge from the mean as time progresses toward the final time. This behavior can be explained by the fact that the affine controller gains tend to increase as the final time approaches. These increasing gains result in more reactive control in the EPM, compensating for errors in the IPM position and velocity trajectories. As a result, the IPM position errors are effectively compensated by the final time, allowing the system to successfully reach the goal positions.

\section{CONCLUSIONS}

This work introduces a novel method for trajectory optimization and control in robotic magnetic manipulation, utilizing a single permanent magnet attached to a robotic arm. The proposed approach leverages a constrained iLQR method to optimize trajectories for both the EPM and IPM while accounting for their respective dynamics and constraints. By employing an augmented Lagrangian framework, the method integrates position and velocity limits for the IPM, magnetic force constraints, and feasible, singularity-free paths for the EPM. The resulting optimized state-input trajectories and time-varying controller gains enable real-time, closed-loop velocity control of the robotic arm, improving the manipulability of the EPM and enhancing control precision.

The experimental results motivated by capsule endoscopy requirements, validate the effectiveness of the method, demonstrating accurate and repeatable positioning of the IPM with minimal deviations. The developed simulation environment, which incorporates detailed models of the robotic arm and magnetic components, closely replicates the real system and serves as a reliable platform for evaluating the proposed control strategy.

From an application perspective, an exciting future direction would be to 
base the navigation on the real-time visual feedback from an internal camera attached to the IPM. This would require incorporating visual localization and control in an unknown environment, while considering the operator's task-specific knowledge and requirements.

\end{document}

%% file: fig2.pdf_tex
\begingroup%
  \makeatletter%
  \providecommand\color[2][]{%
    \errmessage{(Inkscape) Color is used for the text in Inkscape, but the package 'color.sty' is not loaded}%
    \renewcommand\color[2][]{}%
  }%
  \providecommand\transparent[1]{%
    \errmessage{(Inkscape) Transparency is used (non-zero) for the text in Inkscape, but the package 'transparent.sty' is not loaded}%
    \renewcommand\transparent[1]{}%
  }%
  \providecommand\rotatebox[2]{#2}%
  \newcommand*\fsize{\dimexpr\f@size pt\relax}%
  \newcommand*\lineheight[1]{\fontsize{\fsize}{#1\fsize}\selectfont}%
  \ifx\svgwidth\undefined%
    \setlength{\unitlength}{824.17167123bp}%
    \ifx\svgscale\undefined%
      \relax%
    \else%
      \setlength{\unitlength}{\unitlength * \real{\svgscale}}%
    \fi%
  \else%
    \setlength{\unitlength}{\svgwidth}%
  \fi%
  \global\let\svgwidth\undefined%
  \global\let\svgscale\undefined%
  \makeatother%
  \begin{picture}(1,0.37784725)%
    \lineheight{1}%
    \setlength\tabcolsep{0pt}%
    \put(0,0){\includegraphics[width=\unitlength,page=1]{fig2.pdf}}%
    \put(0.69,0.1754058){\color[rgb]{1,1,1}\makebox(0,0)[lt]{\lineheight{1.25}\smash{\begin{tabular}[t]{l}EPM\end{tabular}}}}%
    \put(0.46667194,0.22206116){\color[rgb]{0,0,0}\makebox(0,0)[lt]{\lineheight{1.25}\smash{\begin{tabular}[t]{l}Joint angles, $\mathbf{q}$\end{tabular}}}}%
    \put(0.395,0.28421946){\color[rgb]{0,0,0}\makebox(0,0)[lt]{\lineheight{1.25}\smash{\begin{tabular}[t]{l}$\mathbf{x}_e$\end{tabular}}}}%

    \put(0.58361954,0.28119383){\color[rgb]{0,0,0}\makebox(0,0)[lt]{\lineheight{1.25}\smash{\begin{tabular}[t]{l}$\mathbf{u}$\end{tabular}}}}%

    \put(0.337,0.17){\color[rgb]{0,0,0}\makebox(0,0)[lt]{\lineheight{1.25}\smash{\begin{tabular}[t]{l} $\hat{\mathbf{p}}_I$ $\hat{\mathbf{v}}_I$\end{tabular}}}}%

    \put(0.7207,0.115){\color[rgb]{1,1,1}\makebox(0,0)[lt]{\lineheight{1.25}\smash{\begin{tabular}[t]{l}IPM\end{tabular}}}}%
    \put(0.64555976,0.08649093){\color[rgb]{1,1,1}\rotatebox{-6.00000076}{\makebox(0,0)[lt]{\lineheight{1.25}\smash{\begin{tabular}[t]{l}Camera\end{tabular}}}}}%
    \put(0.80471077,0.08573371){\color[rgb]{1,1,1}\rotatebox{31.19999728}{\makebox(0,0)[lt]{\lineheight{1.25}\smash{\begin{tabular}[t]{l}Camera\end{tabular}}}}}%
    \put(0,0){\includegraphics[width=\unitlength,page=2]{fig2.pdf}}%
    \put(0.705,0.34569547){\color[rgb]{1,1,1}\makebox(0,0)[lt]{\lineheight{1.25}\smash{\begin{tabular}[t]{l}$5$\end{tabular}}}}%
    \put(0.69775461,0.29368418){\color[rgb]{1,1,1}\makebox(0,0)[lt]{\lineheight{1.25}\smash{\begin{tabular}[t]{l}$12$\end{tabular}}}}%
    \put(0.65894607,0.315){\color[rgb]{1,1,1}\rotatebox{90}{\makebox(0,0)[lt]{\lineheight{1.25}\smash{\begin{tabular}[t]{l}$\diameter 6$\end{tabular}}}}}%
    \put(0,0){\includegraphics[width=\unitlength,page=3]{fig2.pdf}}%
    \put(0.72476984,0.33){\color[rgb]{1,1,1}\rotatebox{-90}{\makebox(0,0)[lt]{\lineheight{1.25}\smash{\begin{tabular}[t]{l}$5$\end{tabular}}}}}%
    \put(0.336,0.23){\color[rgb]{0,0,0}\makebox(0,0)[lt]{\lineheight{1.25}\smash{\begin{tabular}[t]{l}$\hat{\mathbf{x}}$\end{tabular}}}}%
    \put(0,0){\includegraphics[width=\unitlength,page=4]{fig2.pdf}}%
    \put(0.04,0.28567996){\color[rgb]{0.0,0.0,0.0}\transparent{1}\makebox(0,0)[lt]{\lineheight{1.25}\smash{\begin{tabular}[t]{c}
    User defined \\
    constraints
    \end{tabular}}}}%
    \put(0.47,0.13){\color[rgb]{0.0,0.0,0.0}\transparent{1}\makebox(0,0)[lt]{\lineheight{1.25}\smash{\begin{tabular}[t]{c}
    Object\\Detection\\
    and\\Triangulation
    \end{tabular}}}}%
    \put(0.075,0.33711995){\color[rgb]{0,0,0}\transparent{1}\makebox(0,0)[lt]{\lineheight{1.25}\smash{\begin{tabular}[t]{l}iLQR \end{tabular}}}}
    \put(0.50673453,0.26881656){\color[rgb]{0,0,0}\makebox(0,0)[lt]{\lineheight{1.25}\smash{\begin{tabular}[t]{l}+\end{tabular}}}}%
    \put(0.35412379,0.25362682){\color[rgb]{0,0,0}\makebox(0,0)[lt]{\lineheight{1.25}\smash{\begin{tabular}[t]{l}-\end{tabular}}}}%
    \put(0.09661046,0.01476934){\color[rgb]{0,0.03529412,0.21568627}\transparent{0.5}\makebox(0,0)[lt]{\lineheight{1.25}\smash{\begin{tabular}[t]{l}Planning\end{tabular}}}}%
    \put(0.90684014,0.01476811){\color[rgb]{0,0.03529412,0.21568627}\transparent{0.5}\makebox(0,0)[lt]{\lineheight{1.25}\smash{\begin{tabular}[t]{l}Control\end{tabular}}}}%
    \put(0,0){\includegraphics[width=\unitlength,page=5]{fig2.pdf}}%
    \put(0.03,0.22564937){\color[rgb]{0.05,0.0,0.0}\transparent{1}\makebox(0,0)[lt]{\lineheight{1.25}\smash{\begin{tabular}[t]{c}
    Initial and Goal\\
    Position
    \end{tabular}}}}%
    \put(0.31,0.11){\color[rgb]{0.0,0.0,0.0}\transparent{1}\makebox(0,0)[lt]{\lineheight{1.25}\smash{\begin{tabular}[t]{c}
    Extended\\ Kalman Filter
    \end{tabular}}}}%

    \put(0,0){\includegraphics[width=\unitlength,page=6]{fig2.pdf}}%
    \put(0.3,0.35341516){\color[rgb]{0,0,0}\makebox(0,0)[lt]{\lineheight{1.25}\smash{\begin{tabular}[t]{l}
    Optimal Input Trajectories $\mathbf{u}^*$
    \end{tabular}}}}%

    \put(0.26671337,0.32559047){\color[rgb]{0,0,0}\makebox(0,0)[lt]{\lineheight{1.25}\smash{\begin{tabular}[t]{l}
     Controller Gains $\mathbf{K}(k)$
    \end{tabular}}}}%

    \put(0.205,0.282){\color[rgb]{0,0,0}\makebox(0,0)[lt]{\lineheight{1.25}\smash{\begin{tabular}[t]{l}
    Optimal State\\
    Trajectories $\mathbf{x}^*$
    \end{tabular}}}}%

    \put(0.52567028,0.28852628){\color[rgb]{0,0,0}\makebox(0,0)[lt]{\lineheight{1.25}\smash{\begin{tabular}[t]{l}+\end{tabular}}}}%
    \put(0,0){\includegraphics[width=\unitlength,page=7]{fig2.pdf}}%
  \end{picture}%
\endgroup%

%% file: fig5.pdf_tex
\begingroup%
  \makeatletter%
  \providecommand\color[2][]{%
    \errmessage{(Inkscape) Color is used for the text in Inkscape, but the package 'color.sty' is not loaded}%
    \renewcommand\color[2][]{}%
  }%
  \providecommand\transparent[1]{%
    \errmessage{(Inkscape) Transparency is used (non-zero) for the text in Inkscape, but the package 'transparent.sty' is not loaded}%
    \renewcommand\transparent[1]{}%
  }%
  \providecommand\rotatebox[2]{#2}%
  \newcommand*\fsize{\dimexpr\f@size pt\relax}%
  \newcommand*\lineheight[1]{\fontsize{\fsize}{#1\fsize}\selectfont}%
  \ifx\svgwidth\undefined%
    \setlength{\unitlength}{1655.78826664bp}%
    \ifx\svgscale\undefined%
      \relax%
    \else%
      \setlength{\unitlength}{\unitlength * \real{\svgscale}}%
    \fi%
  \else%
    \setlength{\unitlength}{\svgwidth}%
  \fi%
  \global\let\svgwidth\undefined%
  \global\let\svgscale\undefined%
  \makeatother%
  \begin{picture}(1,0.35426785)%
    \lineheight{1}%
    \setlength\tabcolsep{0pt}%
    \put(0,0){\includegraphics[width=\unitlength,page=1]{fig5.pdf}}%
    \put(0.03083154,0.22566081){\makebox(0,0)[lt]{\lineheight{1.25}\smash{\begin{tabular}[t]{l}0\end{tabular}}}}%
    \put(0.0670137,0.22566081){\makebox(0,0)[lt]{\lineheight{1.25}\smash{\begin{tabular}[t]{l}1\end{tabular}}}}%
    \put(0.10319134,0.22566081){\makebox(0,0)[lt]{\lineheight{1.25}\smash{\begin{tabular}[t]{l}2\end{tabular}}}}%
    \put(0.1393735,0.22566081){\makebox(0,0)[lt]{\lineheight{1.25}\smash{\begin{tabular}[t]{l}3\end{tabular}}}}%
    \put(0.17555113,0.22566081){\makebox(0,0)[lt]{\lineheight{1.25}\smash{\begin{tabular}[t]{l}4\end{tabular}}}}%
    \put(0.21173329,0.22566081){\makebox(0,0)[lt]{\lineheight{1.25}\smash{\begin{tabular}[t]{l}5\end{tabular}}}}%
    \put(0.24791093,0.22566081){\makebox(0,0)[lt]{\lineheight{1.25}\smash{\begin{tabular}[t]{l}6\end{tabular}}}}%
    \put(0.28409309,0.22566081){\makebox(0,0)[lt]{\lineheight{1.25}\smash{\begin{tabular}[t]{l}7\end{tabular}}}}%
    \put(0.32027072,0.22566081){\makebox(0,0)[lt]{\lineheight{1.25}\smash{\begin{tabular}[t]{l}8\end{tabular}}}}%
    \put(0.01962087,0.23573755){\makebox(0,0)[lt]{\lineheight{1.25}\smash{\begin{tabular}[t]{l}46\end{tabular}}}}%
    \put(0.01962087,0.26054144){\makebox(0,0)[lt]{\lineheight{1.25}\smash{\begin{tabular}[t]{l}48\end{tabular}}}}%
    \put(0.01962087,0.28534081){\makebox(0,0)[lt]{\lineheight{1.25}\smash{\begin{tabular}[t]{l}50\end{tabular}}}}%
    \put(0.01962087,0.31014471){\makebox(0,0)[lt]{\lineheight{1.25}\smash{\begin{tabular}[t]{l}52\end{tabular}}}}%
    \put(0,0){\includegraphics[width=\unitlength,page=2]{fig5.pdf}}%
    \put(0.03083154,0.12335303){\makebox(0,0)[lt]{\lineheight{1.25}\smash{\begin{tabular}[t]{l}0\end{tabular}}}}%
    \put(0.0670137,0.12335303){\makebox(0,0)[lt]{\lineheight{1.25}\smash{\begin{tabular}[t]{l}1\end{tabular}}}}%
    \put(0.10319134,0.12335303){\makebox(0,0)[lt]{\lineheight{1.25}\smash{\begin{tabular}[t]{l}2\end{tabular}}}}%
    \put(0.1393735,0.12335303){\makebox(0,0)[lt]{\lineheight{1.25}\smash{\begin{tabular}[t]{l}3\end{tabular}}}}%
    \put(0.17555113,0.12335303){\makebox(0,0)[lt]{\lineheight{1.25}\smash{\begin{tabular}[t]{l}4\end{tabular}}}}%
    \put(0.21173329,0.12335303){\makebox(0,0)[lt]{\lineheight{1.25}\smash{\begin{tabular}[t]{l}5\end{tabular}}}}%
    \put(0.24791093,0.12335303){\makebox(0,0)[lt]{\lineheight{1.25}\smash{\begin{tabular}[t]{l}6\end{tabular}}}}%
    \put(0.28409309,0.12335303){\makebox(0,0)[lt]{\lineheight{1.25}\smash{\begin{tabular}[t]{l}7\end{tabular}}}}%
    \put(0.32027072,0.12335303){\makebox(0,0)[lt]{\lineheight{1.25}\smash{\begin{tabular}[t]{l}8\end{tabular}}}}%
    \put(0.01889654,0.13342981){\makebox(0,0)[lt]{\lineheight{1.25}\smash{\begin{tabular}[t]{l}-7\end{tabular}}}}%
    \put(0.01889654,0.16319358){\makebox(0,0)[lt]{\lineheight{1.25}\smash{\begin{tabular}[t]{l}-3\end{tabular}}}}%
    \put(0.02528283,0.19295735){\makebox(0,0)[lt]{\lineheight{1.25}\smash{\begin{tabular}[t]{l}1\end{tabular}}}}%
    \put(0,0){\includegraphics[width=\unitlength,page=3]{fig5.pdf}}%
    \put(0.03083154,0.02104525){\makebox(0,0)[lt]{\lineheight{1.25}\smash{\begin{tabular}[t]{l}0\end{tabular}}}}%
    \put(0.0670137,0.02104525){\makebox(0,0)[lt]{\lineheight{1.25}\smash{\begin{tabular}[t]{l}1\end{tabular}}}}%
    \put(0.10319134,0.02104525){\makebox(0,0)[lt]{\lineheight{1.25}\smash{\begin{tabular}[t]{l}2\end{tabular}}}}%
    \put(0.1393735,0.02104525){\makebox(0,0)[lt]{\lineheight{1.25}\smash{\begin{tabular}[t]{l}3\end{tabular}}}}%
    \put(0.17555113,0.02104525){\makebox(0,0)[lt]{\lineheight{1.25}\smash{\begin{tabular}[t]{l}4\end{tabular}}}}%
    \put(0.21173329,0.02104525){\makebox(0,0)[lt]{\lineheight{1.25}\smash{\begin{tabular}[t]{l}5\end{tabular}}}}%
    \put(0.24791093,0.02104525){\makebox(0,0)[lt]{\lineheight{1.25}\smash{\begin{tabular}[t]{l}6\end{tabular}}}}%
    \put(0.28409309,0.02104525){\makebox(0,0)[lt]{\lineheight{1.25}\smash{\begin{tabular}[t]{l}7\end{tabular}}}}%
    \put(0.32027072,0.02104525){\makebox(0,0)[lt]{\lineheight{1.25}\smash{\begin{tabular}[t]{l}8\end{tabular}}}}%
    \put(0.02528283,0.03112203){\makebox(0,0)[lt]{\lineheight{1.25}\smash{\begin{tabular}[t]{l}0\end{tabular}}}}%
    \put(0.02528283,0.0608858){\makebox(0,0)[lt]{\lineheight{1.25}\smash{\begin{tabular}[t]{l}4\end{tabular}}}}%
    \put(0.02528283,0.09064957){\makebox(0,0)[lt]{\lineheight{1.25}\smash{\begin{tabular}[t]{l}8\end{tabular}}}}%
    \put(0.05101516,0.32580341){\makebox(0,0)[lt]{\lineheight{1.25}\smash{\begin{tabular}[t]{l}Goal Position\end{tabular}}}}%
    \put(0,0){\includegraphics[width=\unitlength,page=4]{fig5.pdf}}%
    \put(0.16199782,0.32580341){\makebox(0,0)[lt]{\lineheight{1.25}\smash{\begin{tabular}[t]{l}Mean ± StdDev\end{tabular}}}}%
    \put(0,0){\includegraphics[width=\unitlength,page=5]{fig5.pdf}}%
    \put(0.27870489,0.32580341){\makebox(0,0)[lt]{\lineheight{1.25}\smash{\begin{tabular}[t]{l}Mean pos\end{tabular}}}}%
    \put(0,0){\includegraphics[width=\unitlength,page=6]{fig5.pdf}}%
    \put(0.155,0.34671976){\makebox(0,0)[lt]{\lineheight{1.25}\smash{\begin{tabular}[t]{l}a) IPM Position\end{tabular}}}}%
    \put(0.15541469,0.00270051){\makebox(0,0)[lt]{\lineheight{1.25}\smash{\begin{tabular}[t]{l}Time [s]\end{tabular}}}}%
    \put(0.00920685,0.25747436){\rotatebox{90}{\makebox(0,0)[lt]{\lineheight{1.25}\smash{\begin{tabular}[t]{l}X [cm]\end{tabular}}}}}%
    \put(0.00920287,0.1551666){\rotatebox{90}{\makebox(0,0)[lt]{\lineheight{1.25}\smash{\begin{tabular}[t]{l}Y [cm]\end{tabular}}}}}%
    \put(0.00920596,0.05361524){\rotatebox{90}{\makebox(0,0)[lt]{\lineheight{1.25}\smash{\begin{tabular}[t]{l}Z [cm]\end{tabular}}}}}%
    \put(0.48,0.34671976){\makebox(0,0)[lt]{\lineheight{1.25}\smash{\begin{tabular}[t]{l}b) IPM Velocity\end{tabular}}}}%
    \put(0.80,0.34671976){\makebox(0,0)[lt]{\lineheight{1.25}\smash{\begin{tabular}[t]{l}c) EPM Position\end{tabular}}}}%
    \put(0,0){\includegraphics[width=\unitlength,page=7]{fig5.pdf}}%
    \put(0.36683249,0.22566081){\makebox(0,0)[lt]{\lineheight{1.25}\smash{\begin{tabular}[t]{l}0\end{tabular}}}}%
    \put(0.40301465,0.22566081){\makebox(0,0)[lt]{\lineheight{1.25}\smash{\begin{tabular}[t]{l}1\end{tabular}}}}%
    \put(0.43919228,0.22566081){\makebox(0,0)[lt]{\lineheight{1.25}\smash{\begin{tabular}[t]{l}2\end{tabular}}}}%
    \put(0.47537445,0.22566081){\makebox(0,0)[lt]{\lineheight{1.25}\smash{\begin{tabular}[t]{l}3\end{tabular}}}}%
    \put(0.51155208,0.22566081){\makebox(0,0)[lt]{\lineheight{1.25}\smash{\begin{tabular}[t]{l}4\end{tabular}}}}%
    \put(0.54773424,0.22566081){\makebox(0,0)[lt]{\lineheight{1.25}\smash{\begin{tabular}[t]{l}5\end{tabular}}}}%
    \put(0.58391188,0.22566081){\makebox(0,0)[lt]{\lineheight{1.25}\smash{\begin{tabular}[t]{l}6\end{tabular}}}}%
    \put(0.62009404,0.22566081){\makebox(0,0)[lt]{\lineheight{1.25}\smash{\begin{tabular}[t]{l}7\end{tabular}}}}%
    \put(0.65627167,0.22566081){\makebox(0,0)[lt]{\lineheight{1.25}\smash{\begin{tabular}[t]{l}8\end{tabular}}}}%
    \put(0.35489749,0.25603){\makebox(0,0)[lt]{\lineheight{1.25}\smash{\begin{tabular}[t]{l}-1\end{tabular}}}}%
    \put(0.36128377,0.28308962){\makebox(0,0)[lt]{\lineheight{1.25}\smash{\begin{tabular}[t]{l}1\end{tabular}}}}%
    \put(0.36128377,0.31014471){\makebox(0,0)[lt]{\lineheight{1.25}\smash{\begin{tabular}[t]{l}3\end{tabular}}}}%
    \put(0,0){\includegraphics[width=\unitlength,page=8]{fig5.pdf}}%
    \put(0.36683249,0.12335303){\makebox(0,0)[lt]{\lineheight{1.25}\smash{\begin{tabular}[t]{l}0\end{tabular}}}}%
    \put(0.40301465,0.12335303){\makebox(0,0)[lt]{\lineheight{1.25}\smash{\begin{tabular}[t]{l}1\end{tabular}}}}%
    \put(0.43919228,0.12335303){\makebox(0,0)[lt]{\lineheight{1.25}\smash{\begin{tabular}[t]{l}2\end{tabular}}}}%
    \put(0.47537445,0.12335303){\makebox(0,0)[lt]{\lineheight{1.25}\smash{\begin{tabular}[t]{l}3\end{tabular}}}}%
    \put(0.51155208,0.12335303){\makebox(0,0)[lt]{\lineheight{1.25}\smash{\begin{tabular}[t]{l}4\end{tabular}}}}%
    \put(0.54773424,0.12335303){\makebox(0,0)[lt]{\lineheight{1.25}\smash{\begin{tabular}[t]{l}5\end{tabular}}}}%
    \put(0.58391188,0.12335303){\makebox(0,0)[lt]{\lineheight{1.25}\smash{\begin{tabular}[t]{l}6\end{tabular}}}}%
    \put(0.62009404,0.12335303){\makebox(0,0)[lt]{\lineheight{1.25}\smash{\begin{tabular}[t]{l}7\end{tabular}}}}%
    \put(0.65627167,0.12335303){\makebox(0,0)[lt]{\lineheight{1.25}\smash{\begin{tabular}[t]{l}8\end{tabular}}}}%
    \put(0.35489749,0.14019698){\makebox(0,0)[lt]{\lineheight{1.25}\smash{\begin{tabular}[t]{l}-2\end{tabular}}}}%
    \put(0.36128377,0.16725207){\makebox(0,0)[lt]{\lineheight{1.25}\smash{\begin{tabular}[t]{l}0\end{tabular}}}}%
    \put(0.36128377,0.19430716){\makebox(0,0)[lt]{\lineheight{1.25}\smash{\begin{tabular}[t]{l}2\end{tabular}}}}%
    \put(0,0){\includegraphics[width=\unitlength,page=9]{fig5.pdf}}%
    \put(0.36683249,0.02104525){\makebox(0,0)[lt]{\lineheight{1.25}\smash{\begin{tabular}[t]{l}0\end{tabular}}}}%
    \put(0.40301465,0.02104525){\makebox(0,0)[lt]{\lineheight{1.25}\smash{\begin{tabular}[t]{l}1\end{tabular}}}}%
    \put(0.43919228,0.02104525){\makebox(0,0)[lt]{\lineheight{1.25}\smash{\begin{tabular}[t]{l}2\end{tabular}}}}%
    \put(0.47537445,0.02104525){\makebox(0,0)[lt]{\lineheight{1.25}\smash{\begin{tabular}[t]{l}3\end{tabular}}}}%
    \put(0.51155208,0.02104525){\makebox(0,0)[lt]{\lineheight{1.25}\smash{\begin{tabular}[t]{l}4\end{tabular}}}}%
    \put(0.54773424,0.02104525){\makebox(0,0)[lt]{\lineheight{1.25}\smash{\begin{tabular}[t]{l}5\end{tabular}}}}%
    \put(0.58391188,0.02104525){\makebox(0,0)[lt]{\lineheight{1.25}\smash{\begin{tabular}[t]{l}6\end{tabular}}}}%
    \put(0.62009404,0.02104525){\makebox(0,0)[lt]{\lineheight{1.25}\smash{\begin{tabular}[t]{l}7\end{tabular}}}}%
    \put(0.65627167,0.02104525){\makebox(0,0)[lt]{\lineheight{1.25}\smash{\begin{tabular}[t]{l}8\end{tabular}}}}%
    \put(0.35489749,0.05141448){\makebox(0,0)[lt]{\lineheight{1.25}\smash{\begin{tabular}[t]{l}-1\end{tabular}}}}%
    \put(0.36128377,0.0784741){\makebox(0,0)[lt]{\lineheight{1.25}\smash{\begin{tabular}[t]{l}1\end{tabular}}}}%
    \put(0.36128377,0.10552919){\makebox(0,0)[lt]{\lineheight{1.25}\smash{\begin{tabular}[t]{l}3\end{tabular}}}}%
    \put(0.38959355,0.32580341){\makebox(0,0)[lt]{\lineheight{1.25}\smash{\begin{tabular}[t]{l}Constraint\end{tabular}}}}%
    \put(0,0){\includegraphics[width=\unitlength,page=10]{fig5.pdf}}%
    \put(0.47983097,0.32580341){\makebox(0,0)[lt]{\lineheight{1.25}\smash{\begin{tabular}[t]{l}Mean ± StdDev\end{tabular}}}}%
    \put(0,0){\includegraphics[width=\unitlength,page=11]{fig5.pdf}}%
    \put(0.59110256,0.32580341){\makebox(0,0)[lt]{\lineheight{1.25}\smash{\begin{tabular}[t]{l}Mean velocity\end{tabular}}}}%
    \put(0,0){\includegraphics[width=\unitlength,page=12]{fig5.pdf}}%
    \put(0.49141564,0.00270051){\makebox(0,0)[lt]{\lineheight{1.25}\smash{\begin{tabular}[t]{l}Time [s]\end{tabular}}}}%
    \put(0.3452078,0.25747436){\rotatebox{90}{\makebox(0,0)[lt]{\lineheight{1.25}\smash{\begin{tabular}[t]{l}X [cm/s]\end{tabular}}}}}%
    \put(0.34520382,0.1551666){\rotatebox{90}{\makebox(0,0)[lt]{\lineheight{1.25}\smash{\begin{tabular}[t]{l}Y [cm/s]\end{tabular}}}}}%
    \put(0.34520691,0.05361524){\rotatebox{90}{\makebox(0,0)[lt]{\lineheight{1.25}\smash{\begin{tabular}[t]{l}Z [cm/s]\end{tabular}}}}}%
    \put(0,0){\includegraphics[width=\unitlength,page=13]{fig5.pdf}}%
    \put(0.7055513,0.22566081){\makebox(0,0)[lt]{\lineheight{1.25}\smash{\begin{tabular}[t]{l}0\end{tabular}}}}%
    \put(0.74173346,0.22566081){\makebox(0,0)[lt]{\lineheight{1.25}\smash{\begin{tabular}[t]{l}1\end{tabular}}}}%
    \put(0.7779111,0.22566081){\makebox(0,0)[lt]{\lineheight{1.25}\smash{\begin{tabular}[t]{l}2\end{tabular}}}}%
    \put(0.81409326,0.22566081){\makebox(0,0)[lt]{\lineheight{1.25}\smash{\begin{tabular}[t]{l}3\end{tabular}}}}%
    \put(0.85027089,0.22566081){\makebox(0,0)[lt]{\lineheight{1.25}\smash{\begin{tabular}[t]{l}4\end{tabular}}}}%
    \put(0.88645305,0.22566081){\makebox(0,0)[lt]{\lineheight{1.25}\smash{\begin{tabular}[t]{l}5\end{tabular}}}}%
    \put(0.92263069,0.22566081){\makebox(0,0)[lt]{\lineheight{1.25}\smash{\begin{tabular}[t]{l}6\end{tabular}}}}%
    \put(0.95881285,0.22566081){\makebox(0,0)[lt]{\lineheight{1.25}\smash{\begin{tabular}[t]{l}7\end{tabular}}}}%
    \put(0.99499048,0.22566081){\makebox(0,0)[lt]{\lineheight{1.25}\smash{\begin{tabular}[t]{l}8\end{tabular}}}}%
    \put(0.69434063,0.25437218){\makebox(0,0)[lt]{\lineheight{1.25}\smash{\begin{tabular}[t]{l}49\end{tabular}}}}%
    \put(0.69434063,0.27367718){\makebox(0,0)[lt]{\lineheight{1.25}\smash{\begin{tabular}[t]{l}50\end{tabular}}}}%
    \put(0.69434063,0.29298219){\makebox(0,0)[lt]{\lineheight{1.25}\smash{\begin{tabular}[t]{l}51\end{tabular}}}}%
    \put(0,0){\includegraphics[width=\unitlength,page=14]{fig5.pdf}}%
    \put(0.7055513,0.12335303){\makebox(0,0)[lt]{\lineheight{1.25}\smash{\begin{tabular}[t]{l}0\end{tabular}}}}%
    \put(0.74397107,0.12335303){\makebox(0,0)[lt]{\lineheight{1.25}\smash{\begin{tabular}[t]{l}1\end{tabular}}}}%
    \put(0.78238631,0.12335303){\makebox(0,0)[lt]{\lineheight{1.25}\smash{\begin{tabular}[t]{l}2\end{tabular}}}}%
    \put(0.82080607,0.12335303){\makebox(0,0)[lt]{\lineheight{1.25}\smash{\begin{tabular}[t]{l}3\end{tabular}}}}%
    \put(0.85922131,0.12335303){\makebox(0,0)[lt]{\lineheight{1.25}\smash{\begin{tabular}[t]{l}4\end{tabular}}}}%
    \put(0.89764108,0.12335303){\makebox(0,0)[lt]{\lineheight{1.25}\smash{\begin{tabular}[t]{l}5\end{tabular}}}}%
    \put(0.93605632,0.12335303){\makebox(0,0)[lt]{\lineheight{1.25}\smash{\begin{tabular}[t]{l}6\end{tabular}}}}%
    \put(0.97447609,0.12335303){\makebox(0,0)[lt]{\lineheight{1.25}\smash{\begin{tabular}[t]{l}7\end{tabular}}}}%
    \put(0.6936163,0.14226246){\makebox(0,0)[lt]{\lineheight{1.25}\smash{\begin{tabular}[t]{l}-5\end{tabular}}}}%
    \put(0.70000259,0.16775938){\makebox(0,0)[lt]{\lineheight{1.25}\smash{\begin{tabular}[t]{l}0\end{tabular}}}}%
    \put(0.70000259,0.19325177){\makebox(0,0)[lt]{\lineheight{1.25}\smash{\begin{tabular}[t]{l}5\end{tabular}}}}%
    \put(0,0){\includegraphics[width=\unitlength,page=15]{fig5.pdf}}%
    \put(0.7055513,0.02104525){\makebox(0,0)[lt]{\lineheight{1.25}\smash{\begin{tabular}[t]{l}0\end{tabular}}}}%
    \put(0.74397107,0.02104525){\makebox(0,0)[lt]{\lineheight{1.25}\smash{\begin{tabular}[t]{l}1\end{tabular}}}}%
    \put(0.78238631,0.02104525){\makebox(0,0)[lt]{\lineheight{1.25}\smash{\begin{tabular}[t]{l}2\end{tabular}}}}%
    \put(0.82080607,0.02104525){\makebox(0,0)[lt]{\lineheight{1.25}\smash{\begin{tabular}[t]{l}3\end{tabular}}}}%
    \put(0.85922131,0.02104525){\makebox(0,0)[lt]{\lineheight{1.25}\smash{\begin{tabular}[t]{l}4\end{tabular}}}}%
    \put(0.89764108,0.02104525){\makebox(0,0)[lt]{\lineheight{1.25}\smash{\begin{tabular}[t]{l}5\end{tabular}}}}%
    \put(0.93605632,0.02104525){\makebox(0,0)[lt]{\lineheight{1.25}\smash{\begin{tabular}[t]{l}6\end{tabular}}}}%
    \put(0.97447609,0.02104525){\makebox(0,0)[lt]{\lineheight{1.25}\smash{\begin{tabular}[t]{l}7\end{tabular}}}}%
    \put(0.69434063,0.03484534){\makebox(0,0)[lt]{\lineheight{1.25}\smash{\begin{tabular}[t]{l}20\end{tabular}}}}%
    \put(0.69434063,0.05375627){\makebox(0,0)[lt]{\lineheight{1.25}\smash{\begin{tabular}[t]{l}25\end{tabular}}}}%
    \put(0.69434063,0.0726672){\makebox(0,0)[lt]{\lineheight{1.25}\smash{\begin{tabular}[t]{l}30\end{tabular}}}}%
    \put(0.69434063,0.09158266){\makebox(0,0)[lt]{\lineheight{1.25}\smash{\begin{tabular}[t]{l}35\end{tabular}}}}%
    \put(0.72604758,0.32580341){\makebox(0,0)[lt]{\lineheight{1.25}\smash{\begin{tabular}[t]{l}Constraint\end{tabular}}}}%
    \put(0,0){\includegraphics[width=\unitlength,page=16]{fig5.pdf}}%
    \put(0.81356726,0.32580341){\makebox(0,0)[lt]{\lineheight{1.25}\smash{\begin{tabular}[t]{l}Mean ± StdDev\end{tabular}}}}%
    \put(0,0){\includegraphics[width=\unitlength,page=17]{fig5.pdf}}%
    \put(0.92574476,0.32580341){\makebox(0,0)[lt]{\lineheight{1.25}\smash{\begin{tabular}[t]{l}Mean pos\end{tabular}}}}%
    \put(0,0){\includegraphics[width=\unitlength,page=18]{fig5.pdf}}%
    \put(0.83013446,0.00270051){\makebox(0,0)[lt]{\lineheight{1.25}\smash{\begin{tabular}[t]{l}Time [s]\end{tabular}}}}%
    \put(0.68392661,0.25747436){\rotatebox{90}{\makebox(0,0)[lt]{\lineheight{1.25}\smash{\begin{tabular}[t]{l}X [cm]\end{tabular}}}}}%
    \put(0.68392263,0.1551666){\rotatebox{90}{\makebox(0,0)[lt]{\lineheight{1.25}\smash{\begin{tabular}[t]{l}Y [cm]\end{tabular}}}}}%
    \put(0.68392572,0.05361524){\rotatebox{90}{\makebox(0,0)[lt]{\lineheight{1.25}\smash{\begin{tabular}[t]{l}Z [cm]\end{tabular}}}}}%
  \end{picture}%
\endgroup%

%% file: fig6.pdf_tex
\begingroup%
  \makeatletter%
  \providecommand\color[2][]{%
    \errmessage{(Inkscape) Color is used for the text in Inkscape, but the package 'color.sty' is not loaded}%
    \renewcommand\color[2][]{}%
  }%
  \providecommand\transparent[1]{%
    \errmessage{(Inkscape) Transparency is used (non-zero) for the text in Inkscape, but the package 'transparent.sty' is not loaded}%
    \renewcommand\transparent[1]{}%
  }%
  \providecommand\rotatebox[2]{#2}%
  \newcommand*\fsize{\dimexpr\f@size pt\relax}%
  \newcommand*\lineheight[1]{\fontsize{\fsize}{#1\fsize}\selectfont}%
  \ifx\svgwidth\undefined%
    \setlength{\unitlength}{1940.99991926bp}%
    \ifx\svgscale\undefined%
      \relax%
    \else%
      \setlength{\unitlength}{\unitlength * \real{\svgscale}}%
    \fi%
  \else%
    \setlength{\unitlength}{\svgwidth}%
  \fi%
  \global\let\svgwidth\undefined%
  \global\let\svgscale\undefined%
  \makeatother%
  \begin{picture}(1,0.68624422)%
    \lineheight{1}%
    \setlength\tabcolsep{0pt}%
    \put(0,0){\includegraphics[width=\unitlength,page=1]{fig6.pdf}}%
    \put(0.19390409,0.42612129){\color[rgb]{0,0,0}\makebox(0,0)[lt]{\lineheight{1.25}\smash{\begin{tabular}[t]{l}Goal\end{tabular}}}}%
    \put(0,0){\includegraphics[width=\unitlength,page=2]{fig6.pdf}}%
    \put(0.9549859,0.09303534){\color[rgb]{0,0,0}\makebox(0,0)[t]{\lineheight{1.25}\smash{\begin{tabular}[t]{c}\text{x}\end{tabular}}}}%
    \put(0.90046143,0.10774573){\color[rgb]{0,0,0}\makebox(0,0)[t]{\lineheight{1.25}\smash{\begin{tabular}[t]{c}\text{z}\end{tabular}}}}%
    \put(0.81407649,0.05292701){\color[rgb]{0,0,0}\makebox(0,0)[t]{\lineheight{1.25}\smash{\begin{tabular}[t]{c}\text{y}\end{tabular}}}}%
    \put(0.88028196,0.17869125){\color[rgb]{0,0,0}\makebox(0,0)[t]{\lineheight{1.25}\smash{\begin{tabular}[t]{c}Start\end{tabular}}}}%
  \end{picture}%
\endgroup%